\documentclass{article}

\usepackage[usenames,dvipsnames,table]{xcolor}
\usepackage{iclr2026_conference,times}
\usepackage[T1]{fontenc}    %
\usepackage[colorlinks=true,linkcolor=MidnightBlue,urlcolor=MidnightBlue,citecolor=MidnightBlue]{hyperref}
\usepackage{url}            %
\usepackage{booktabs}       %
\usepackage{amsmath}        %
\usepackage{amssymb}        %
\usepackage{mathtools}      %
\usepackage{bm}             %
\usepackage{microtype}      %
\usepackage{nicefrac}       %
\usepackage{pifont}         %
\usepackage{siunitx}        %
\usepackage{xspace}         %
\usepackage{enumitem}       %
\usepackage{cleveref}       %
\usepackage{environ}        %
\usepackage{subcaption}
\usepackage{graphicx}
\usepackage{listings}
\usepackage{wrapfig}
\usepackage[nohypertypes={acronym,main}]{glossaries}
\usepackage[normalem]{ulem}
\definecolor{lightgray}{gray}{0.95}
\robustify\bfseries
\robustify\uline

\usepackage{acro}

\iclrfinalcopy

\newcommand{\autocite}[1]{\citep{#1}}
\newcommand{\textcite}[1]{\citet{#1}}

\newif\ifshowdraft
\showdrafttrue  %

\NewEnviron{draft}{%
  \ifshowdraft
  {\color{red}\BODY}%
  \fi
}

\newlist{inlineroman}{enumerate*}{1}
\setlist[inlineroman]{afterlabel=~,label=(\roman*)}
\newlist{inlinearabic}{enumerate*}{1}
\setlist[inlinearabic]{afterlabel=~,label=\arabic*)}

\providecommand{\given}{\,\vert\,}
\DeclarePairedDelimiterX{\parens}[1]{\lparen}{\rparen}{\renewcommand\given{\nonscript\:\delimsize\vert\nonscript\:\mathopen{}}#1}
\newcommand{\p}[1][p]{{#1}\parens}

\renewcommand{\vec}[1]{\bm{#1}}
\newcommand{\mat}[1]{\bm{\mathrm{#1}}}

\DeclareMathOperator{\dv}{\mathrm{d}\!}

\DeclarePairedDelimiterXPP{\bigO}[1]{\operatorname{\mathcal{O}}}{\lparen}{\rparen}{}{\renewcommand\given{\nonscript\:\delimsize\vert\nonscript\:\mathopen{}}#1}

\newcommand\norm[1]{\left\lVert#1\right\rVert}

\DeclarePairedDelimiterXPP{\expect}[1]{\operatorname{\mathbb{E}}}{[}{]}{}{\renewcommand\given{\nonscript\:\delimsize\vert\nonscript\:\mathopen{}}#1}

\DeclarePairedDelimiterXPP{\var}[1]{\operatorname{Var}}{[}{]}{}{\renewcommand\given{\nonscript\:\delimsize\vert\nonscript\:\mathopen{}}#1}

\DeclarePairedDelimiterXPP{\entropy}[1]{\operatorname{H}}{[}{]}{}{\renewcommand\given{\nonscript\:\delimsize\vert\nonscript\:\mathopen{}}#1}

\DeclarePairedDelimiterXPP{\mode}[1]{\operatorname{mode}}{[}{]}{}{\renewcommand\given{\nonscript\:\delimsize\vert\nonscript\:\mathopen{}}#1}

\newcommand{\distribution}[1]{{#1}\parens*}
\newcommand{\gaussian}{\distribution{\operatorname{\mathcal{N}}}}

\newcommand{\gammadist}{\distribution{\operatorname{\mathrm{Gam}}}}
\newcommand{\invgammadist}{\distribution{\operatorname{\mathrm{InvGam}}}}
\newcommand{\studentt}{\distribution{\operatorname{\mathrm{St}}}}
\newcommand{\chisq}[2]{\distribution{\operatorname{\chi}^2_{#1}}{#2}}

\DeclareMathOperator{\snr}{SNR}
\DeclareMathOperator{\snri}{SNRi}
\DeclareMathOperator{\snrref}{SNR_{ref}}

\definecolor{codegreen}{rgb}{0,0.6,0}
\definecolor{codegray}{rgb}{0.5,0.5,0.5}
\definecolor{codepurple}{rgb}{0.58,0,0.82}
\definecolor{backcolour}{rgb}{0.95,0.95,0.92}

\lstdefinestyle{mystyle}{
backgroundcolor=\color{backcolour},
commentstyle=\color{codegreen},
keywordstyle=\color{magenta},
numberstyle=\tiny\color{codegray},
stringstyle=\color{codepurple},
basicstyle=\ttfamily\footnotesize,
breakatwhitespace=false,
breaklines=true,
captionpos=b,
keepspaces=true,
numbers=left,
numbersep=5pt,
showspaces=false,
showstringspaces=false,
showtabs=false,
tabsize=2
}

\DeclareAcronym{snr}{short = SNR, long = signal-to-noise ratio}
\DeclareAcronym{snri}{short = SNRi, long = \ac{snr} improvement}
\DeclareAcronym{sdr}{short = SDR, long = signal-to-distortion ratio}
\DeclareAcronym{sdri}{short = SDRi, long = \ac{sdr} improvement}
\DeclareAcronym{sisnr}{short = SI-SNR, long = scale-invariant signal-to-noise ratio}
\DeclareAcronym{sisnri}{short = SI-SNRi, long = \ac{sisnr} improvement}
\DeclareAcronym{sisdr}{short = SI-SDR, long = scale-invariant
signal-to-distortion ratio}
\DeclareAcronym{sisdri}{short = SI-SDRi, long = \ac{sisdr} improvement}
\DeclareAcronym{upit}{short = uPIT, long = utterance-level permutation
invariant training}
\DeclareAcronym{tasnet}{short = TasNet, long = Time-domain Audio Separation Network}
\DeclareAcronym{this}{short = PRESS, long = PRobabilistic Early-exit for
Speech Separation}
\DeclareAcronym{glu}{short = GLU, long = gated linear unit}
\DeclareAcronym{geglu}{short = GeGLU, long = GELU gated linear unit}
\DeclareAcronym{rnn}{short = RNN, long = recurrent neural network}
\DeclareAcronym{sota}{short = SOTA, long = state-of-the-art}
\DeclareAcronym{rms}{short = RMS, long = root mean square}
\DeclareAcronym{db}{short = dB, long = decibel}
\DeclareAcronym{dbfs}{
  short = dBFS,
  long = \ac{db} relative to full scale,
  short-plural-form = dBFS,
  long-plural-form = \acp{db} relative to full scale
}
\DeclareAcronym{gmac}{short = GMAC, long = giga-multiply-accumulates}
\DeclareAcronym{gmacs}{short = GMAC/s, long = \ac{gmac} per second}
\DeclareAcronym{ece}{short = ECE, long = expected calibration error}
\DeclareAcronym{pit}{short = PIT, long = probability integral transform}
\DeclareAcronym{pce}{short = PCE, long = probabilistic calibration error}
\DeclareAcronym{crps}{short = CRPS, long = continuous ranked probability score}
\DeclareAcronym{ks}{short = KS, long = Kolmogorov-Smirnov}

\makeglossaries

\showdrafttrue

\title{Knowing When to Quit: Probabilistic Early Exits for Speech Separation Networks}

\author{%
  Kenny Falkær Olsen$^{1,2}$ \quad Mads Østergaard$^{2}$ \quad Karl
  Ulbæk$^{2}$ \quad Søren Føns Nielsen$^{2}$\\
  \bf \quad Rasmus Malik Høegh Lindrup$^2$ \quad Bjørn Sand Jensen$^1$ \quad
  Morten Mørup$^1$\\[2ex]
  $^1$Technical University of Denmark \quad $^2$WS Audiology
}

\begin{document}

\maketitle

\begin{abstract}
  In recent years, deep learning-based single-channel speech separation has improved
  considerably, in large part driven by increasingly compute- and parameter-efficient
  neural network architectures. Most such architectures are, however, designed with a
  fixed compute and parameter budget and consequently cannot scale to varying compute
  demands or resources, which limits their use in embedded and heterogeneous
  devices such as mobile phones and hearables.
  To enable such use-cases we design a neural network architecture for speech separation
  and enhancement capable of early-exit, and we propose an uncertainty-aware
  probabilistic framework to jointly model the clean speech signal and error variance
  which we use to derive probabilistic early-exit conditions in terms of desired
  signal-to-noise ratios.
  We evaluate our methods on both speech separation and enhancement tasks where we
  demonstrate that early-exit capabilities can be introduced without compromising
  reconstruction, and that when trained on variable-length audio our early-exit
  conditions are well-calibrated and lead to considerable compute savings when used to
  dynamically scale compute at test time while remaining directly interpretable.
\end{abstract}

\section{Introduction}

The cocktail party problem~\autocite{CherrySomeExperimentsOn1953} concerns the
separation of a (possibly unknown) number of overlapping speakers, potentially
corrupted by environmental noise and reverberation. The task is typically divided into
speech separation (separating multiple speakers) and speech enhancement (removing
environmental noise and/or reverberation), which have many applications in, e.g.,
telecommunications and hearables.

Single-channel speech separation has become predominantly deep learning-based with the
introduction of the \ac{tasnet}~\autocite{LuoTasnet2018}. Notable follow-ups include
Conv-\ac{tasnet}~\autocite{LuoConvTasnet2019}, which demonstrated competitive speech
separation performance could be achieved at significantly lower computational cost than
previously held, and SepFormer~\autocite{SubakanAttentionIsAll2021}, which showed large
performance improvements using a transformer-based architecture. While the transformer
improved modeling performance and training parallelism over prior recurrent and
convolutional networks it also tied computational complexity to the length of the
attention context window, which may become prohibitive for long sequence or in online
contexts.
For training, these works have primarily relied on the \ac{sisnr}
loss~\autocite{RouxSdrHalfBaked2018} with
\ac{upit}~\autocite{YuPermutationInvariantTraining2017,KolbaekMultiTalkerSpeech2017}.

\begin{figure}[!t]
  \centering
  \includegraphics[width=\linewidth]{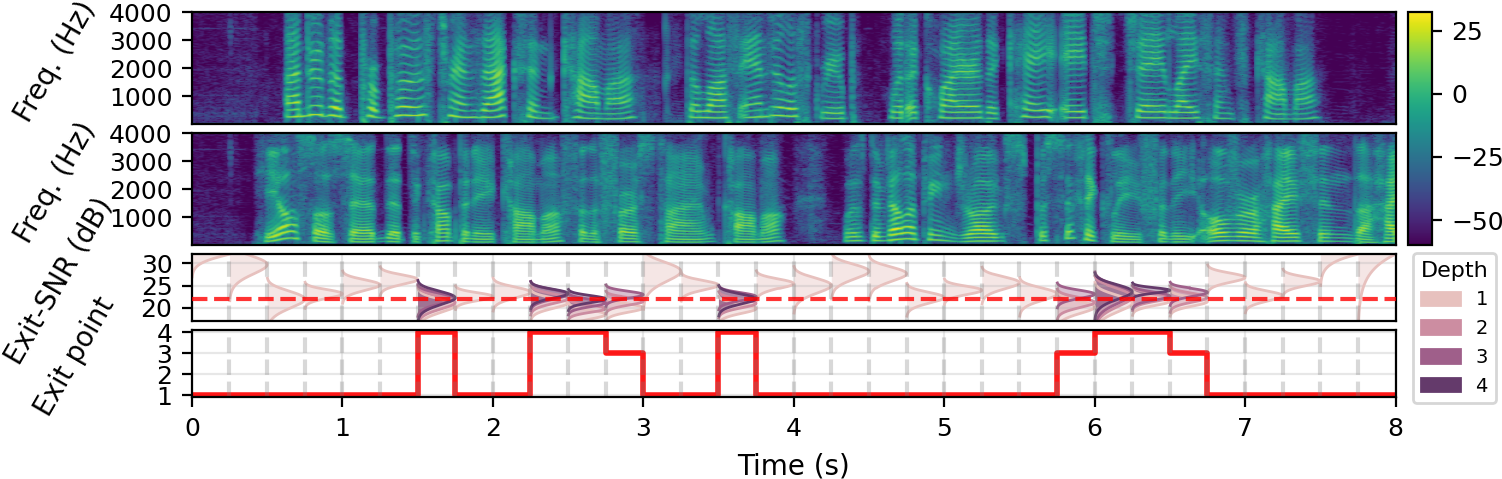}
  \caption{
    \textbf{Reconstructed spectrograms} of two speakers from the WSJ0-2mix test
    set separated by a PRESS-4-S model with 4 exit points evaluated in segments
    of $T=2000$ samples, showing our proposed exit-SNR exit condition evaluated for
    each segment with a target level of $22$ dB (shown in \textcolor{red}{red}). The
    distributions of each exit-SNR condition is shown shaded by exit point,
    demonstrating non-trivial improvement for deeper exits. An extended version
    showing our other SNR-like distributions can be seen in~\cref{sec:full_demo},
    \cref{fig:early_exit_demo_full}.
  } \label{fig:early_exit_demo}
\end{figure}

Current \ac{sota} speech separation and enhancement architectures typically cannot vary
their compute in response to simplifying conditions in the input such as
non-overlapping speech, low environmental noise, or silence. Following the terminology
in~\textcite{MontelloASurveyOn2025}, we refer to neural networks which \emph{can} vary
compute based on either self-estimated difficulty or external conditions as
\textit{dynamic}, while networks that cannot are \textit{static}. One method for
introducing dynamism in neural networks is \textit{early
exit}~\autocite{ScardapaneWhyShouldWe2020}, where the network can be used to make
predictions at several depths through the network stack based on an exit condition.
Other methods include \textit{dynamic routing} where various subsets of the network
weights are used during inference based on a routing condition, such as
mixture-of-experts~\autocite{MuAComprehensiveSurvey2025} (MoE) and slimmable
networks~\autocite{YuSlimmableNeuralNetworks2018}.

Early-exit allows direct scaling of both compute and the total number of activated
parameters in the network, yielding both energy and prediction latency
savings~\autocite{JazbecFastYetSafe2024}, while allowing fine-grained decision-making
using all information and processing available up to the current exit point. End-to-end
training of early-exiting neural networks has also been shown to reduce overfitting and
improve gradient propagation during training~\autocite{ScardapaneWhyShouldWe2020}. In
the context of speech separation, a
SepReformer~\autocite{ShinSeparateAndReconstruct2024} trained with additional
reconstruction losses reminiscent of early exit improved the final performance of the
network.

Prior work on early exit typically defines the exit condition implicitly through the
loss function by defining a reconstruction loss and a utilization loss and minimizing a
convex combination of the
two~\autocite{TanSparseUniversalTransformer2023,BraliosLatentIterativeRefinement2023,ElminshawiDynamicSlimmableNetwork2024},
which leads to an implicit performance-compute trade-off that is frozen during
training, and hence cannot be adapted at inference time. Other work instead defines the
exit condition based on similarity between consecutive exit
points~\autocite{ChenDonTShoot2020}, which does not ground the exit condition in any
performance metric (which the first kind does through the loss).

\paragraph{Contributions}
In this work, we introduce \ac{this} which leverages early exits to reduce inference
costs. Importantly, our approach is probabilistic and naturally assign weights to the
different exit conditions by their quantified log-likelihood-based reconstruction
quality. At inference time, the probabilistic formulation enables \ac{this} to
control when to exit computation using the model's confidence that a desired \ac{snr}
between the estimated and clean audio signals has been achieved, providing a
directly interpretable exit condition grounded in the network's performance at
each exit point.

Specifically, we make the following contributions:
\begin{enumerate}[left=2pt]%
  \item We propose a new early-exit framework with uncertainty awareness though a
    probabilistic model formulation accounting for both the clean speech signal and
    the variance of the error as well as its associated uncertainty. We demonstrate
    that our approach provides a simple and principled framework to balance
    optimization of reconstruction quality with early-exit accuracy without the
    need for careful weighting of multiple objectives. Using the probabilistic
    model we construct several fully probabilistic \ac{snr}-like early-exit
    conditions which can be used to early-exit when a target \ac{snr} level has
    been reached with a given uncertainty tolerance.
  \item To instantiate our framework, we propose a new speech separation architecture
    based on linear \acp{rnn} designed with the joint goals of (1) achieving
    \ac{sota}-level reconstruction performance per compute, and (2) being
    architecturally capable of outputting high-quality reconstructions from its
    early-exit points.
  \item We validate our approach for both speech separation on the
    WSJ0-2mix~\autocite{GarofoloCsrIComplete2007},
    Libri2Mix~\autocite{TzinisHeterogeneousTargetSpeech2022},
    WHAM!~\autocite{WichernWham2019}, WHAMR!~\autocite{MaciejewskiWhamr2020}
    datasets, and for speech enhancement on the DNS Challenge
    2020~\autocite{ReddyTheInterspeechDeep2020} dataset, demonstrating the
    viability of training a single, dynamic neural network with early exits to
    achieve performance competitive with \ac{sota} static, single-exit models.
\end{enumerate}

\section{Related Work}

\paragraph{\Ac{tasnet}-family} \Ac{tasnet}~\autocite{LuoTasnet2018} replaced masking in
a fixed spectral representation
with learnable convolutional encoder and decoder layers. The bulk of parameters and
compute of \acp{tasnet} are in a \ac{rnn} masker network, which produces masks that are
applied to the encoded mixture before decoding to produce single-source estimates.
\ac{tasnet} inspired follow-up work such as computationally efficient variants like the
Conv-\ac{tasnet}~\autocite{LuoConvTasnet2019} and
SudoRMRF~\autocite{TzinisSudoRmRf2020}, as well as the
DP-RNN~\autocite{LuoDualPathRnn2020} which introduced dual-path processing,
decoupling time-mixing operations into chunks.
The dual-path structure was also used with transformer-based networks in
SepFormer~\autocite{SubakanAttentionIsAll2021}, TF-GridNet~\autocite{WangTfGridnet2023}
and MossFormer~\autocite{ZhaoMossformer2023,ZhaoCombiningTransformerAnd2023}.

\paragraph{SepReformer} SepReformer~\autocite{ShinSeparateAndReconstruct2024}
introduces \textit{early
splitting}; instead of projecting a shared encoded representation into separate
estimated sources late in the network as in \ac{tasnet}, the projection into separate
sources occurs early in the network. Further processing happens independently for each
source with only a cross-speaker attention layer enabling information exchange between
processing for the sources. SepReformer is constructed as a
U-net~\autocite{RonnebergerUNet2015} with a transformer-like stack
\autocite{VaswaniAttentionIsAll2023} composed of a convolutional block, a time-mixing
attention block, and a cross-speaker attention block, and uses
LayerScale~\autocite{TouvronGoingDeeperWith2021} to train comparatively deep and slim
networks. The network maximizes \ac{sisnr} using \ac{upit} with extra loss terms
maximizing the \ac{sisnr} of downsampled frequency-domain representations of the
estimated sources to encourage early separation of sources.

\paragraph{Diffusion models, SNR estimation and iterative refinement}
SepIt~\autocite{LutatiSeparateAndDiffuse2023} iteratively refines its estimates and
stops processing based on bounding the SNR through mutual information between current
estimates and the input mixture---this bound is extended to consider generative models
in DiffSep~\autocite{ScheiblerDiffusionBasedGenerative2022}. Diffusion/score-based
models like DiffSep can, generally, sample the learned diffusion process with variable
compute requirements in a trade-off with quality. For instance, DiffWave
\autocite{KongDiffwave2021} --- a diffusion-based vocoder also used in separation tasks
in Separate and Diffuse \cite{LutatiSeparateAndDiffuse2023}---shows faster sampling by
reducing needed steps through designing an appropriate variance schedule.

\paragraph{Slimmable Networks}
allow the \emph{width} of the network to be adjusted at inference time to trade-off
compute and accuracy have been termed slimmable
networks~\autocite{YuSlimmableNeuralNetworks2018,YuUniversallySlimmableNetworks2019,LiDynamicSlimmableNetwork2021}.
This can be achieved using switchable batch
normalization~\autocite{YuSlimmableNeuralNetworks2018}, which essentially, given a
predefined set of widths, trains a single network with multiple batch normalization
layers, one pr. width. The width specific batch-normalization can be moved to
post-training~\autocite{YuUniversallySlimmableNetworks2019}, employing knowledge
distillation during training while sampling different widths pr. batch. The idea of
\emph{dynamic} slimmable networks~\autocite{LiDynamicSlimmableNetwork2021} uses a
dynamic gating mechanism that routes the signal to a subset of the next stage in the
network to reduce complexity. In the context of speech and audio,
Slim-\ac{tasnet}~\autocite{ElminshawiSlimTasnet2023} combines the \ac{tasnet}
architecture for
speech separation with width slimming and a given input width size. A dynamic version
of this, dynamic slimmable \ac{tasnet}~\autocite{ElminshawiDynamicSlimmableNetwork2024},
utilizes a predictive scheme, where each subnetwork predicts how much of the following
subnetwork that should be utilized. Similarly, in the context of speech enhancement,
dynamic channel pruning~\autocite{MicciniDynamic2023} is a technique for estimating a
mask applied in the channel dimension of convolutional layers, that reduces the runtime
computational cost.

\paragraph{Early Exit}
Various strategies are used to construct early-exit conditions, and to train early-exit
models. In classification tasks, such as image classification, the entropy at each exit
has been used as an exit
condition~\autocite{TeerapittayanonBranchynet2017,ScardapaneDifferentiableBranchingIn2020}
using pre-softmax logits as a proxy for model uncertainty. In the context of language
modeling the Sparse Universal Transformer~\autocite{TanSparseUniversalTransformer2023}
used a stick-breaking construction to define monotonically increasing halting
probabilities which down-weighted intermediate activations in later layers to scale
down their influence on the final prediction, which was learned through a single loss.
In the context of speech separation,
the Euclidean norm difference between successive
blocks has been used as an exit condition in a transformer-based
architecture~\autocite{ChenDonTShoot2020} and early-exit has also been used
based on a learned gating
function in an iterative model~\autocite{BraliosLatentIterativeRefinement2023}.
In speech enhancement,
PDRE~\autocite{NakataniAHybridProbabilistic2025} iteratively applies a deterministic
U-net network to the noisy signal and predicts the parameters of a Gaussian mixture
model (GMM), to get a distribution over the clean signal at each iteration, and the
network is trained to maximize the weighted sum of log-likelihoods of each step's GMM.
However, no stopping criteria or exit-conditions were explored to determine when to
stop the enhancement process.

\section{Methods}\label{sec:methods}

The goal of single-channel speech separation neural networks (and enhancement as a
special case) is to separate an input mixture signal $\vec{\widetilde{x}} \in
\mathbb{R}^T$ into a set of estimated sources $\vec{\widehat{x}}_i \in \mathbb{R}^T$
which approximate the target sources $\vec{x}_j \in \mathbb{R}^T$ in the input mixture,
where $S$ is the total number of sources and $i,j \in [1, S]$ . To incorporate
early-exit into such networks, we require (1) a set of early-exit \emph{conditions}
which can be used to decide when an estimate is acceptable enough to exit, (2) an
\emph{objective} to learn both the source reconstruction and the exit conditions, and
(3) a neural network \emph{architecture} which can support early exit without
compromising reconstruction performance.

\subsection{Probabilistic Speech Modelling}\label{sec:prob_separation}

The performance of speech separation and enhancement systems is often reported as
\acfp{snr} or \acp{snri} (i.e. the relative gain in \ac{snr} by using
$\vec{\widehat{x}}_i$ over the input signal $\vec{\widetilde{x}}$), computed as the
ratio of the power of the target signal $\vec{x}_j$ to be estimated and the power of
the error signal $\vec{x}_j - \vec{\widehat{x}}_i$,
\begin{equation}
  \snr(\vec{x}_j, \vec{\widehat{x}}_i) = \frac{\norm{\vec{x}_j}_2^2}{\norm{\vec{x}_j
  - \vec{\widehat{x}}_i}_2^2}, \qquad \snri(\vec{x}_j, \vec{\widehat{x}}_i,
  \vec{\widetilde{x}})
  = \frac{\snr(\vec{x}_j, \vec{\widehat{x}}_i)}{{\snr(\vec{x}_j,
  \vec{\widetilde{x}})}}
  = \frac{\norm{\vec{x}_j - \vec{\widetilde{x}}}_2^2}{\norm{\vec{x}_j -
  \vec{\widehat{x}}_i}_2^2}.
\end{equation}
Using \ac{snr} and \ac{snri} as early-exit conditions for speech separation is
desirable because they conceptually measure the loudness of the error relative to the
loudness of the target (or the improvement over the target), ensuring that the system
can optimistically exit when an acceptable speech-to-noise balance is obtained.

Both \ac{snr} and \ac{snri} however require access to the target $\vec{x}_j$, which is
unknown in a predictive setting. We therefore propose to probabilistically model both
the target $\vec{x}_j$ and the error of the prediction using a simple Bayesian
objective where the target signal $\vec{x}_j$ is modelled by jointly predicting an
estimated signal $\vec{\widehat{x}}_i$ and a variance parameter $\sigma_i^2 \in
\mathbb{R}$, where we assume a Gaussian distribution on the signal error and a
conjugate inverse-gamma prior on the variance. Marginalizing out the variance, we
obtain a multivariate Student t-likelihood:
\begin{align}
  \mathcal{L}_i     & {}= \int \gaussian{\vec{x}_j \given \vec{\widehat{x}}_i,
  \sigma_i^2 \mat{I}} \invgammadist{\sigma_i^2 \given \alpha_i, \beta_i} \dv \sigma_i^2
  = \studentt{\vec{x}_j \given \vec{\widehat{x}}_i, 2\alpha_i, \frac{\beta_i}{\alpha_i}
  \mat{I}},\label{eq:studentt_likelihood}                                      \\
  \ln \mathcal{L}_i & \propto \ln \Gamma\parens*{\alpha_i + \frac{T}{2}} - \ln
  \Gamma\parens*{\alpha_i} - \frac{T}{2} \ln \beta_i - \parens*{\alpha_i + \frac{T}{2}}
  \ln\parens*{1 + \frac{\norm{\vec{x}_j - \vec{\widehat{x}}_i}_2^2}{2\beta_i}},
\end{align}
where $\Gamma(\cdot)$ is the gamma function and $\alpha_i$ and $\beta_i$ are the shape
and scale in the inverse-gamma parameterization, which are to be predicted by the model
along with the estimated signal $\vec{\widehat{x}}_i$. This objective strikes a simple
balance between reducing the ratio of the signal error and variance scale through the
last term, while also being penalized for underestimating the variance by the
second-to-last term. See~\cref{sec:scale_invariance} for a scale-invariant version of
this objective and relationship to \ac{sisnr}. When optimizing this objective we use
\ac{upit} to assign targets $\vec{x}_j$ to estimated sources and parameters
$\vec{\widehat{x}}_i, \alpha_i, \beta_i$ by taking the maximum likelihood permutation.
When training with multiple exits, we jointly permute all exits together such that
speakers cannot swap between consecutive exits, and optimize the total likelihood by
summing over all exits and speakers without any weighting.

\paragraph{Predictive Signal-to-Noise Ratios} Using the distributional
assumptions on the target and error signals allows us to construct early exit
conditions expressed directly as predictive signal-to-noise ratios, providing an
interpretable thresholding mechanism for early exit. Based on our model
assumptions, we have
\begin{gather}
  \vec{x}_j \sim \gaussian{\vec{\widehat{x}}_i, \sigma_i^2 \mat{I}},
  \qquad \norm{\vec{x}}_2^2 \sim \sigma_i^2
  \chisq{T}{\frac{\norm{\vec{\widehat{x}}_i}_2^2}{\sigma_i^2}},\\
  \norm{\vec{x}_j - \vec{\widehat{x}}_i}_2^2 \sim \sigma_i^2 \chi^2_T,
  \qquad \norm{\vec{x}_j - \vec{\widetilde{x}}}_2^2 \sim \sigma_i^2
  \chisq{T}{\frac{\norm{\vec{\widehat{x}}_i - \vec{\widetilde{x}}_i}_2^2}{\sigma_i^2}},
\end{gather}
where $\chisq{T}{\lambda}$ is a non-central chi-square distribution with non-centrality
parameter $\lambda$ and $T$ degrees of freedom. The \ac{snr} and \ac{snri} can now be
written as ratios of (non-central) chi-square distributions,
\begin{gather}
  \snr\parens*{\vec{x}_j, \vec{\widehat{x}}_i} =
  \frac{\norm{\vec{x}_j}_2^2}{\norm{\vec{x}_j
  - \vec{\widehat{x}}_i}_2^2} = \frac{\phi_\text{SNR}}{\epsilon},
  \qquad \phi_\text{SNR} \sim
  \chisq{T}{\frac{\norm{\vec{\widehat{x}}_i}_2^2}{\sigma_i^2}},
  \qquad \epsilon \sim \chi^2_T, \\
  \snri\parens*{\vec{x}_j, \vec{\widehat{x}}_i, \vec{\widetilde{x}}} =
  \frac{\norm{\vec{x}_j - \vec{\widetilde{x}}_i}_2^2}{\norm{\vec{x}_j -
  \vec{\widehat{x}}_i}_2^2} = \frac{\phi_\text{SNRi}}{\epsilon},
  \qquad \phi_\text{SNRi} \sim \chisq{T}{\frac{\norm{\vec{\widehat{x}}_i -
  \vec{\widetilde{x}}}_2^2}{\sigma_i^2}}.
\end{gather}
These chi-square distributions are not independent, but the ratio of even dependent
chi-square variables with equal degrees of freedom quickly concentrates around its mean
for large $T$ (see~\cref{sec:chisquare} for details), and so in the limit of large $T$
we approximate the ratios with their conditional means and see that the expressions
takes the form of shifted gamma distributions,
\begin{gather}
  \snr(\vec{x}_j, \vec{\widehat{x}}_i) \xrightarrow{T \to \infty} 1 +
  \frac{\norm{\vec{\widehat{x}}_i}_2^2}{T \sigma_i^2} = 1 + z_\text{SNR}, \qquad
  z_\text{SNR} \sim \gammadist{\alpha_i,
  \frac{\norm{\vec{\widehat{x}}_i}_2^2}{\beta_i T}}, \\
  \snri(\vec{x}_j, \vec{\widehat{x}}_i, \vec{\widetilde{x}}) \xrightarrow{T \to
  \infty} 1 + \frac{\norm{\vec{\widehat{x}}_i - \vec{\widetilde{x}}}_2^2}{T
  \sigma_i^2} = 1 + z_\text{SNRi}, \qquad z_\text{SNRi} \sim \gammadist{\alpha_i,
  \frac{\norm{\vec{\widehat{x}}_i - \vec{\widetilde{x}}}_2^2}{\beta_i T}}.
\end{gather}

\paragraph{Unified Early-Exit SNR} Both \ac{snr} and \ac{snri} can be overly
pessimistic as exit conditions when used in isolation, as silence in the
target such that $\norm{\vec{x}_j}_2^2 \approx 0$ will cause the local \ac{snr} to
approach zero, and a lack of interfering sources such that $\norm{\vec{x}_j -
\vec{\widetilde{x}}}_2^2 \approx 0$ similarly causes the \ac{snri} to vanish.
Additionally, in the case of total silence in all sources both \ac{snr} and
\ac{snri} vanish. To mitigate these silence issues, we propose to use a third
auxiliary loudness condition measuring the \ac{snr} between a fixed
reference signal $\vec{x}_\text{ref} \in \mathbb{R}^T$ with average power
$P_\text{ref}^2 \in \mathbb{R}$ and the predicted noise signal,
\begin{gather}
  \snrref(\vec{x}_j) = \frac{\norm{\vec{x}_\text{ref}}_2^2}{\norm{\vec{x}_j -
  \vec{\widehat{x}}_i}_2^2}
  {}\xrightarrow{T \to \infty} \frac{P_\text{ref}^2}{\sigma_i^2} = z,
  \qquad z \sim \gammadist{\alpha_i, \frac{P_\text{ref}^2}{\beta_i}}.
\end{gather}
This third condition can be used to exit whenever the predicted noise is quieter than
the reference signal by some level difference, where $P_\text{ref}^2$ should be set
based on some tolerable lower bound on the loudness of the noise.

We can now integrate all three exit conditions into a single unified condition, where
we consider the maximum of the individual complementary CDFs for a target \ac{snr}
level $t \in \mathbb{R}$,
\begin{equation}
  \p{\snr_\text{exit}\parens*{\vec{x}_j, \vec{\widehat{x}}_i, \vec{\widetilde{x}}}
  \geq t} = \max \left\{
    \begin{aligned}
      & \p{\snr\parens*{\vec{x}_j, \vec{\widehat{x}}_i} \geq t},
      \\
      & \p{\snri\parens*{\vec{x}_j, \vec{\widehat{x}}_i, \vec{\widetilde{x}}}
      \geq t},                                                                 \\
      & \p{\snrref\parens*{\vec{x}_j} \geq t}
  \end{aligned}\right\},\label{eq:exit_snr}
\end{equation}
which corresponds to \emph{optimistically} exiting when at least one of the conditions
is met with sufficient confidence. We note that $\p{\snr_\text{exit} \geq t}$ is itself
a valid complementary CDF whose full distribution can be given in terms of its
individual component PDFs and CDFs.

To obtain our final combined exit condition across all speakers we take the minimum
over the complementary exit-SNR CDFs for all speakers and exit when the target level
$t$ is exceeded with confidence $p \in [0, 1]$,
\begin{equation}
  \min_i \p{\snr_\text{exit}\parens*{\vec{x}_j, \vec{\widehat{x}}_i,
    \vec{\widetilde{x}}}
  \geq t} \geq p,\label{eq:min_exit_snr}
\end{equation}
which corresponds to \emph{pessimistically} exiting only when all speaker meet at least
one condition with sufficient confidence. We show the predicted distributions of this
condition in~\cref{fig:early_exit_demo}, where $P_\text{ref}^2$ is set to $-35$
\ac{dbfs} such that a target level of $t=22$ \ac{db} corresponds to an estimated noise
loudness below $-57$ \ac{dbfs}, and a full version in~\cref{sec:full_demo},
\cref{fig:early_exit_demo_full} where the individual component distributions are also
shown. We leave the extension of early-exiting each speaker individually as important
future work.

\paragraph{Calibration} Using to our probabilistic modelling setup we can readily
investigate whether our predicted $\sigma^2$ distributions are calibrated ---
that is, they model the actual distribution of the observed mean error power
$\widehat{\sigma}_i^2 = \frac{\norm{\vec{x}_j - \vec{\widehat{x}}_i}_2^2}{T}$ well.
We consider two methods for evaluating this: (1) the
\ac{pit}~\autocite{DheurALargeScale2023} evaluates the predicted CDF on the
observed mean error $\widehat{\sigma}_i^2$ i.e. computing $Z_i =
\p[F_{\sigma_i^2}]{\widehat{\sigma}_i^2}$, which, given perfect calibration, should
itself be uniformly distributed --- we use this property to construct
calibration curves in~\cref{fig:calibration}, and (2) the
\ac{crps}~\autocite{MathesonScoringRulesFor1976} which is a proper scoring rule
generalizing the mean absolute error to distributional regression.

\paragraph{Block-wise Likelihood} Modelling a single, global $\sigma_i^2$ assumes
a stationary error signal which is not realistic for real-world audio. We also
experiment with a blocked variant of the likelihood
in~\cref{eq:studentt_likelihood}, where we model blocks of $T$ consecutive samples
by independent $\sigma_{i,b}^2$ for every block $b$. This redefines $T$ to a fixed
block size rather than the full length of the audio signal. Note that this
likelihood does not require the underlying neural network to be evaluated in
blocks, and only serves to control which samples in the estimated sources are modelled
by the corresponding predicted block-wise inverse-gamma parameters.

\subsection{Model Architecture}\label{sec:arch}

\begin{figure}[t]
  \centering
  \includegraphics[width=\linewidth]{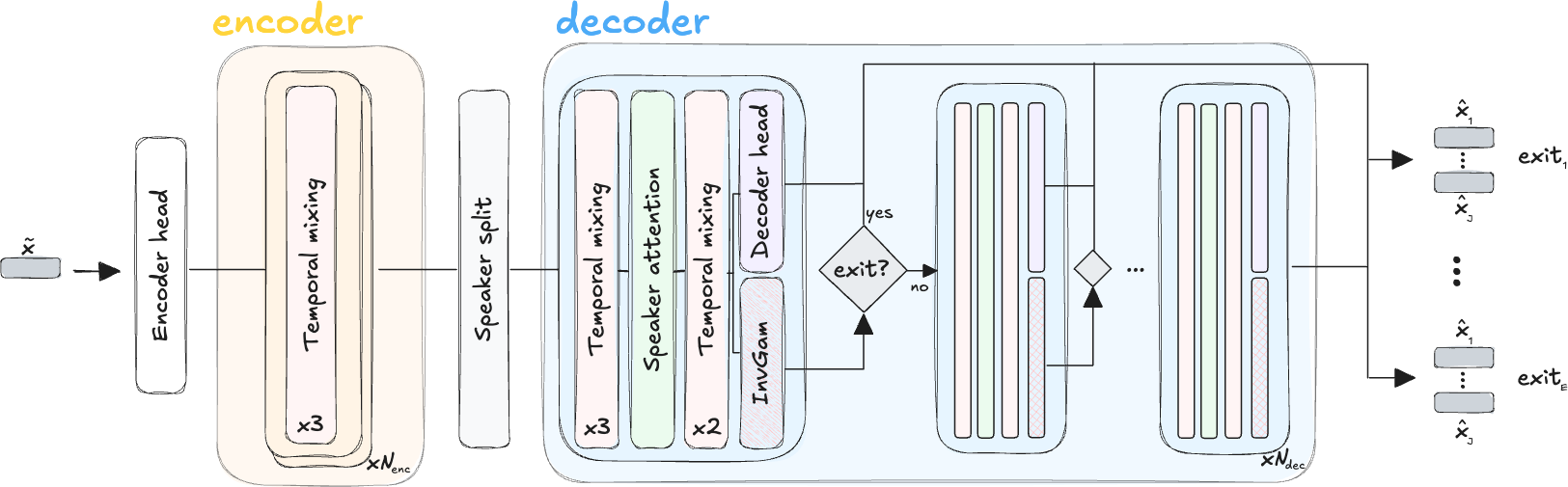}
  \caption{
    \textbf{Detailed architecture} of \ac{this}-Net. It consists of three
    parts: an encoder, an early split module and a reconstruction decoder with
    the ability to reconstruct early.
  } \label{fig:arch:detailed}
\end{figure}

We design a \ac{tasnet}-family model suitable for early exit using \ac{this}, and
denote the architecture \ac{this}-Net as illustrated in~\cref{fig:arch:detailed}. We
base the initial design of the \ac{this}-Net architecture on building blocks from the
SepReformer~\autocite{ShinSeparateAndReconstruct2024} model due to its strong
performance in speech separation. Following recent
trends~\autocite{ShinSeparateAndReconstruct2024,WangTfGridnet2023}, we opt for an
encoder-separator-decoder design where a shallow encoder/decoder pair (referred to as
encoder/decoder heads in our models) down-/upsamples the audio signals whereas a
separator module directly maps the encoder output to the decoder, rather than masking
the encoder output as in \ac{tasnet}. The encoder/decoder heads and inverse-gamma
parametrization blocks are described in~\cref{sec:arch_details}. We do not perform any
down- or upsampling in the separator network in order to be able to reconstruct
estimated sources early in the network without excessive artifacts from upsampling. We
use the GELU~\autocite{HendrycksGaussianErrorLinear2023} activation function.

\paragraph{Separator} We construct our separator module as a deep
transformer-like stack with pre-norm, skip connections, and
LayerScale~\autocite{TouvronGoingDeeperWith2021,WortsmanStableAndLow2023}.
Specifically, for every layer $f$ with input $\vec{x}$ in the stack we compute its
output as $\vec{x} \gets \vec{x} + \vec{\gamma}
\mathop{f}(\operatorname{norm}(\vec{x}))$,
where $\mathrm{norm}(\cdot)$ is an RMSNorm~\autocite{ZhangRootMeanSquare2019}
layer with per-channel scales initialized to
$1$ and $\epsilon=10^{-2}$ as in~\textcite{ParkerScalingTransformersFor2024}, and
$\vec{\gamma} \in \mathbb{R}^D$ is the per-channel LayerScale scaling factors, which we
initialize to $10^{-5}$. This allows very deep networks to be trained stably as the
separator is approximately a skip connection at initialization and only later learns to
integrate non-linearity as $\vec{\gamma}$ grows.

Due to not downsampling in the separator module the time-resolution of the intermediate
activations is much higher than in SepReformer, precluding the direct application of
self-attention over time due to its prohibitively high cost from the quadratic compute
scaling. Instead, we use linear RNNs with self-gating as our primary building block, as
well as speaker attention layers from SepReformer throughout the separator. These
blocks are described in more detail in~\cref{sec:arch_details}, and illustrated
in~\cref{fig:arch:blocks}.

We use \textit{early split} as in SepReformer, where the first $N_{\mathrm{Enc}}$
layers (consisting only of linear RNN blocks) in the stack process the encoded speech
mixture after which a \textit{speaker split} module $\mathrm{SpeakerSplit}:
\mathbb{R}^{T \times D} \to \mathbb{R}^{T \times S \times D}$ projects the refined
speech mixture into $S$ separate groups of channels, where the speaker dimension is
considered a batch dimension in later processing.

After splitting, we repeat a decoder block $N_{\mathrm{Dec}}$ times which consists of
linear RNN blocks and speaker attention blocks repeated in a 5:1 ratio. After each
block, we may place an exit point $E_i < N_{\mathrm{Dec}}$ which reconstructs the
current latent representation with a separate decoder head. Apart from reconstruction,
each exit point parametrizes the inverse gamma distribution which we can use to
evaluate whether we should exit or not.

\section{Results and Discussion}

\begin{table}[t]
  \fontsize{8}{9}\selectfont
  \renewcommand{\arraystretch}{1.2}
  \setlength{\tabcolsep}{3pt}
  \sisetup{
    table-number-alignment = center,
    table-format = 2.2(2),
    detect-all = true,
    separate-uncertainty = true
  }
  \begin{subtable}[c]{.55\linewidth}
    \centering
    \begin{tabular}{
        l
        l
        S[table-format=2.2]
        S[table-format=2.2]
        S[table-format=2.2]
      }
          & {\textbf{Ablation}}               & {\textbf{SI-SNRi}} & {\textbf{SDRi}} & {\textbf{\# Params}} \\
      \toprule
      (a) & SI-SNR loss                       & 22.95              & 23.1            & 3.55M                \\
      (b) & Normal likelihood loss            & 22.42              & 22.58           & 3.55M                \\
      (c) & t-likelihood + per-exit \ac{upit} & 21.1               & 20.97           & 3.55M                \\
      (d) & t-likelihood + 6 exits            & 22.89              & 23.01           & 3.57M                \\
      (e) & t-likelihood + 12 exits           & 22.9               & 22.99           & 3.66M                \\
      (f) & t-likelihood + 200K finetune      & 22.9               & 23.11           & 3.55M                \\
      \bottomrule
    \end{tabular}
    \caption[hmm]{
      Training setup ablations.
    }
  \end{subtable}%
  \begin{subtable}[c]{.45\linewidth}
    \centering
    \begin{tabular}{
        l
        S[table-format=2.2]
        S[table-format=2.2]
        S[table-format=4]
      }
      {\textbf{Ablation}} & {\textbf{SI-SNRi}} & {\textbf{SDRi}} & {\textbf{Receptive field}} \\
      \toprule
      $T=8000$            & 22.82              & 22.98           & 1000 ms                    \\
      $T=4000$            & 22.81              & 22.99           & 500 ms                     \\
      $T=2000$            & 22.79              & 22.96           & 250 ms                     \\
      $T=1000$            & 22.69              & 22.91           & 125 ms                     \\
      $T=500$             & 22.69              & 22.86           & 62 ms                      \\
      \bottomrule
    \end{tabular}
    \caption[hmm]{
      Block size ablations.
    }
  \end{subtable}
  \caption[hmm]{
    \textbf{Ablation results} on the WSJ0-2mix test set.
  }\label{table:result_ablations}
\end{table}

\begin{table}[p]
  \fontsize{8}{9}\selectfont
  \renewcommand{\arraystretch}{1.1}
  \setlength{\tabcolsep}{3pt}
  \newcommand{\hdr}[2]{{\selectfont\shortstack{\textbf{#1}\\{\fontsize{7}{8}\selectfont{#2}}}}}
  \begin{tabular}{
      l
      S[table-format=2.1]
      S[table-format=2.1]
      S[table-format=2.1]
      S[table-format=2.1]
      S[table-format=2.1]
      S[table-format=2.1]
      S[table-format=2.1]
      S[table-format=2.1]
      S[table-format=2.1]
      S[table-format=3.1]
    }
                                 & \multicolumn{2}{c}{\textbf{WSJ0-2mix}} & \multicolumn{2}{c}{\textbf{Libri2Mix}} & \multicolumn{2}{c}{\textbf{WHAM!}} & \multicolumn{2}{c}{\textbf{WHAMR!}} &                     &                                                                                                                  \\
    \cmidrule(lr){2-3} \cmidrule(lr){4-5} \cmidrule(lr){6-7} \cmidrule(lr){8-9}
    \hdr{Model}{}                & \hdr{SI-SNRi}{(dB)}                    & \hdr{SDRi}{(dB)}                       & \hdr{SI-SNRi}{(dB)}                & \hdr{SDRi}{(dB)}                    & \hdr{SI-SNRi}{(dB)} & \hdr{SDRi}{(dB)} & \hdr{SI-SNRi}{(dB)} & \hdr{SDRi}{(dB)} & \hdr{\# Params}{(M)} & \hdr{GMAC/s}{(G/s)}           \\
    \toprule
    {Conv-\ac{tasnet}$^\dagger$} & 15.3                                   & 15.6                                   & 12.2                               & 12.7                                & 12.7                & \textendash      & 8.3                 & \textendash      & 5.1                  & 10.5                          \\
    {DualPathRNN}                & 18.8                                   & 19.0                                   & 16.1                               & 16.6                                & 13.7                & 14.1             & 10.3                & \textendash      & 2.6                  & 42.5                          \\
    \midrule
    {SepFormer}                  & 20.4                                   & 20.5                                   & 19.2                               & 19.4                                & 14.7                & 16.8             & 14.0                & \textendash      & 26.0                 & 86.9                          \\
    {SepFormer + DM$^\dagger$}   & 22.3                                   & 22.5                                   & \textendash                        & \textendash                         & 16.4                & 16.7             & 14.0                & 13               & 26.0                 & 86.9                          \\
    \midrule
    {MossFormer (S)}             & 20.9                                   & \textendash                            & \textendash                        & \textendash                         & \textendash         & \textendash      & \textendash         & \textendash      & 10.8                 & {\textendash\footnotemark[2]} \\
    {MossFormer (M) + DM}        & 22.5                                   & \textendash                            & \textendash                        & \textendash                         & \textendash         & \textendash      & \textendash         & \textendash      & 25.3                 & {\textendash\footnotemark[2]} \\
    {MossFormer (L) + DM}        & 22.8                                   & \textendash                            & \textendash                        & \textendash                         & 17.3                & \textendash      & 16.3                & \textendash      & 42.1                 & 70.4                          \\
    {MossFormer2 + DM}           & 24.1                                   & \textendash                            & 21.7                               & \textendash                         & 18.1                & \textendash      & 17.0                & \textendash      & 55.7                 & 84.2                          \\
    \midrule
    {TF-GridNet (S)}             & 20.6                                   & \textendash                            & \textendash                        & \textendash                         & \textendash         & \textendash      & \textendash         & \textendash      & 8.2                  & 19.2                          \\
    {TF-GridNet (M)}             & 22.2                                   & \textendash                            & \textendash                        & \textendash                         & \textendash         & \textendash      & \textendash         & \textendash      & 8.4                  & 36.2                          \\
    {TF-GridNet (L)}             & 23.4                                   & 23.5                                   & \textendash                        & \textendash                         & \textendash         & \textendash      & 17.3                & 15.8             & 14.4                 & 231.1                         \\
    \midrule
    {SepMamba (S) + DM}          & 21.2                                   & 21.4                                   & \textendash                        & \textendash                         & \textendash         & \textendash      & \textendash         & \textendash      & 7.2                  & 12.5                          \\
    {SepMamba (M) + DM}          & 22.7                                   & 22.9                                   & \textendash                        & \textendash                         & \textendash         & \textendash      & \textendash         & \textendash      & 22.0                 & 37.0                          \\
    \midrule
    {SepReformer (T)}            & 22.4                                   & 22.6                                   & 19.7                               & 20.2                                & 17.2                & 17.5             & \textendash         & \textendash      & 3.7                  & 10.4                          \\
    {SepReformer (S)}            & 23.0                                   & 23.1                                   & 20.6                               & 21.0                                & 17.3                & 17.7             & \textendash         & \textendash      & 4.5                  & 21.3                          \\
    {SepReformer (M)}            & 24.2                                   & 24.4                                   & 22.0                               & 22.2                                & 17.8                & 18.1             & \textendash         & \textendash      & 17.3                 & 81.3                          \\
    {SepReformer (L) + DM}       & 25.1                                   & 25.2                                   & \textendash                        & \textendash                         & 18.4                & 18.7             & 17.2                & 16.0             & 55.3                 & 155.5                         \\
    \midrule
    \rowcolor{lightgray}
    {PRESS-4 @ 4 (S)}            & 22.91                                  & 23.08                                  & 20.04                              & 20.41                               & 16.49               & 16.91            & 14.54               & 13.37            & 3.4                  & 11.3                          \\
    \rowcolor{lightgray}
    {PRESS-12 @ 4 (M)}           & 22.64                                  & 22.93                                  & 19.75                              & 19.71                               & 16.43               & 16.71            & 14.24               & 13.09            & 8.7                  & 29.1                          \\
    \rowcolor{lightgray}
    {PRESS-12 @ 8 (M)}           & 23.47                                  & 24                                     & 20.42                              & 20.86                               & 16.57               & 17.03            & 14.67               & 13.45            & 15.6                 & 54.4                          \\
    \rowcolor{lightgray}
    {PRESS-12 @ 12 (M)}          & 24.28                                  & 24.46                                  & 20.88                              & 21.31                               & 16.65               & 17.12            & 14.69               & 13.47            & 22.4                 & 79.7                          \\
    \midrule
    \rowcolor{lightgray}
    {PRESS-4 @ 4 (S) + FT}       & 23.41                                  & 23.56                                  & 21.01                              & 21.36                               & 17.25               & 17.58            & 15.13               & 13.92            & 3.4                  & 11.3                          \\
    \rowcolor{lightgray}
    {PRESS-12 @ 4 (M) + FT}      & 23.27                                  & 23.43                                  & 20.31                              & 20.72                               & 16.47               & 16.81            & 14.91               & 13.73            & 8.7                  & 29.1                          \\
    \rowcolor{lightgray}
    {PRESS-12 @ 8 (M) + FT}      & 24.18                                  & 24.40                                  & 20.92                              & 21.33                               & 17.21               & 17.34            & 15.61               & 14.38            & 15.6                 & 54.4                          \\
    \rowcolor{lightgray}
    {PRESS-12 @ 12 (M) + FT}     & 24.36                                  & 24.55                                  & 21.29                              & 21.68                               & 17.49               & 17.89            & 15.67               & 14.43            & 22.4                 & 79.7                          \\
    \bottomrule
  \end{tabular}
  \centering
  \caption[hmm]{
    \textbf{Speech separation performance} on the WSJ0-2Mix, Libri2Mix, WHAM! and
    WHAMR! 2-speaker test sets with our PRESS variants highlighted with a gray
    background. We evaluate our PRESS-4 (S) model on its final exit
    point, and also show results for our larger PRESS-12 (M) model at 3 exit
    points. +DM: models trained with dynamic mixing. +FT: models finetuned on
    full-length training data. $^\dagger$: some values
    from~\cite{ShinSeparateAndReconstruct2024}.
  }\label{table:result_comparison}
\end{table}

\begin{figure}[p]
  \centering
  \includegraphics[width=\linewidth]{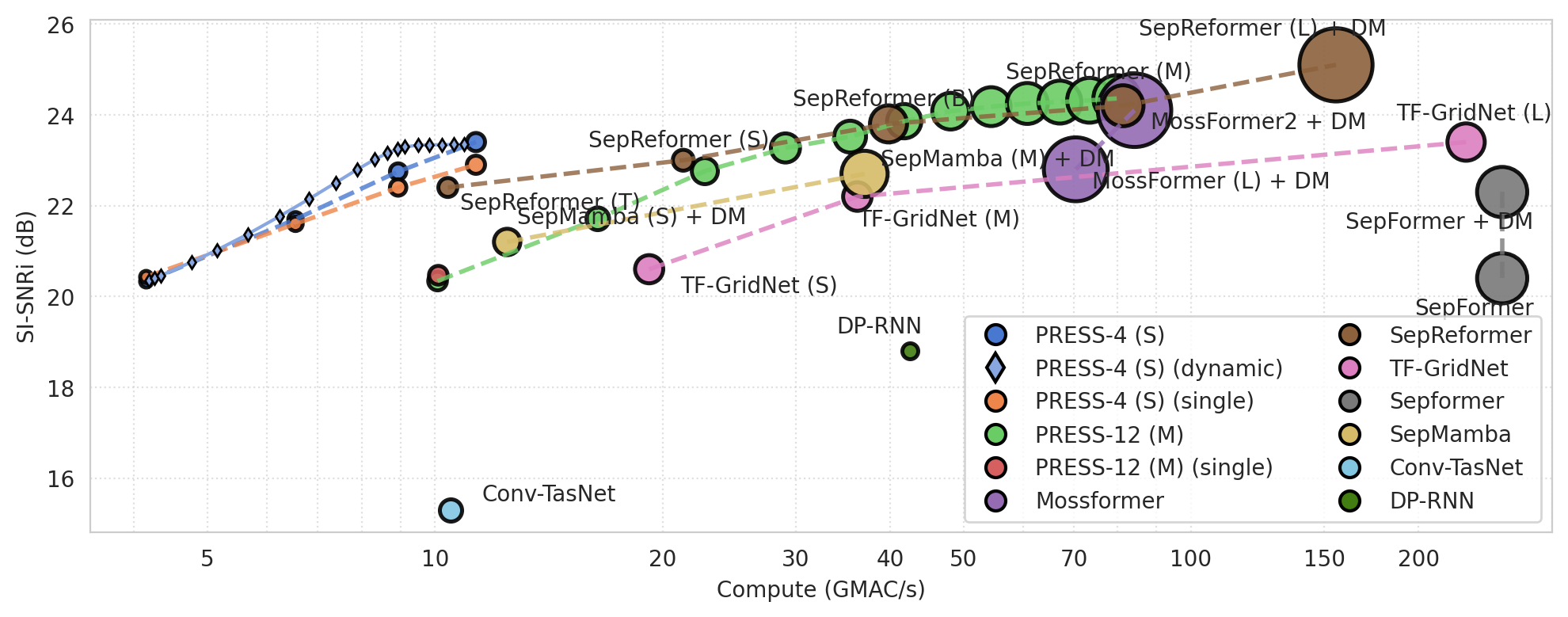}
  \caption[hmm]{
    \textbf{Source separation performance} on WSJ0-2mix in terms of \ac{sisnri} per
    compute (\ac{gmacs}), with the area of points corresponding to parameter
    count of models. The static performance of every exit point is shown for
    \ac{this} models, as well as the dynamic performance of the PRESS-4 (S) model
    using our probabilistic exit condition for varying target levels, beating
    the static performance curve in efficiency. We also include the performance
    of single-exit models, which underperform the jointly trained model at deeper exits.
  } \label{fig:compute_pareto}
\end{figure}

\Ac{this} is evaluated on four speech separation corpora:
WSJ0-2mix~\autocite{GarofoloCsrIComplete2007},
Libri2Mix~\autocite{CosentinoLibrimix2020}, WHAM!~\autocite{WichernWham2019},
and WHAMR!~\autocite{MaciejewskiWhamr2020}, as well as one denoising corpora:
DNS2020~\autocite{ReddyTheInterspeechDeep2020}. Complete descriptions of the
corpora can be found in~\cref{sec:dataset_descriptions}.

We train two main model configurations: (1) PRESS-4 (S), a smaller model with model
dimension $D=64$, $P=4$, $D_\mathrm{Enc}=256$, $N_\mathrm{Enc}=8$, $N_\mathrm{Dec}=12$
and 4 exit points placed at every 3 decoder blocks, and (2) PRESS-12 (M), a larger
model with model dimension increased to $D=128$, $N_\mathrm{Enc}=4$,
$N_\mathrm{Dec}=24$ and having 12 exits, placed at every second decoder block. Further
details of model training can be found in~\cref{sec:training_details}.

We show the performance of our models in~\cref{table:result_comparison} compared with
other \ac{sota} methods, and in~\cref{fig:compute_pareto} we plot performance in
\ac{sisnri} as a function of \ac{gmacs}, showing how \ac{this} models can scale their
compute dynamically while remaining competitive on WSJ0-2Mix.

\begin{figure}[t]
  \centering
  \begin{subfigure}[t]{0.24\linewidth}
    \centering
    \includegraphics[width=1.05\linewidth]{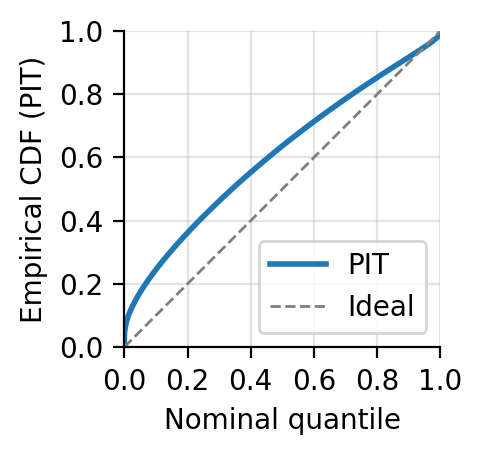}
    \caption[hmm]{
      Full-length training \\data. \\$\text{CRPS} = 1.61$
    }
  \end{subfigure}
  \begin{subfigure}[t]{0.24\linewidth}
    \centering
    \includegraphics[width=1.05\linewidth]{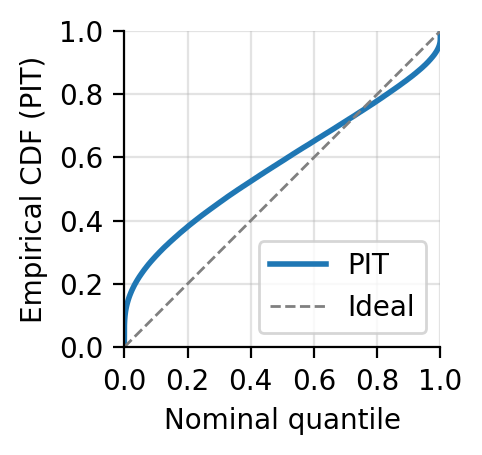}
    \caption[hmm]{
      Full-length test \\data. \\$\text{CRPS} = 2.96$
    }
  \end{subfigure}
  \begin{subfigure}[t]{0.24\linewidth}
    \centering
    \includegraphics[width=1.05\linewidth]{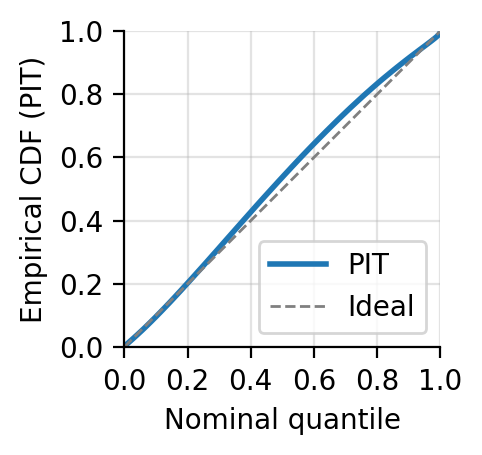}
    \caption[hmm]{
      Full-length training \\data --- \textbf{finetuned}. \\$\text{CRPS} = 1.43$
    }
  \end{subfigure}
  \begin{subfigure}[t]{0.24\linewidth}
    \centering
    \includegraphics[width=1.05\linewidth]{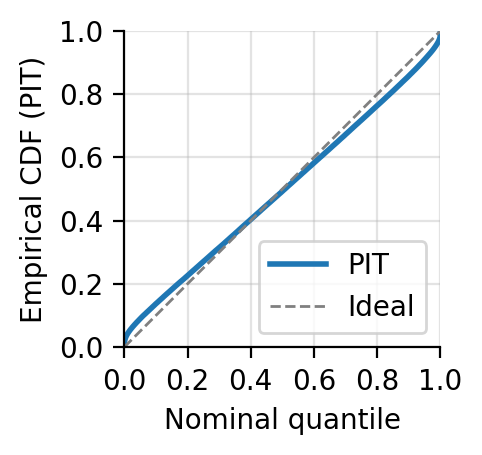}
    \caption[hmm]{
      Full-length test \\data --- \textbf{finetuned}. \\$\text{CRPS} = 2.80$
    }
  \end{subfigure}
  \caption[hmm]{
    \textbf{Calibration curves} for the predicted $\sigma_i^2$ mean error
    distributions on the WSJ0-2mix test set for a PRESS-4 (S) model with a
    block size of 2000 samples. In \textbf{(a)} and \textbf{(b)} we see that
    the distributions are uncalibrated when the model is trained on 4-second
    clips and evaluated on full-length sequences on both training and test
    data. In \textbf{(c)} and \textbf{(d)} we see that the model predictions
    become well-calibrated on both training and test data after finetuning on
    full-length training data.
  }
  \label{fig:calibration}
\end{figure}

In~\cref{fig:calibration} we plot calibration curves for the WSJ0-2Mix training and
test sets, showing that our models are not well-calibrated after training on 4-second
audio clips, and experiment with finetuning these with just 200K extra training steps
(ca. $3\%$ of base training time) on full-length audio clips from the training set,
after which we see our models become well-calibrated, and performance also increases
substantially which we show in~\cref{table:result_comparison} in the bottom rows.

We train several ablation variants of PRESS-4 (S) to determine the effectiveness of (a)
using \ac{sisnr} as loss instead of our t-likelihood, (b) using a normal likelihood
with a single predicted variance instead of a normal-inverse-gamma likelihood (c) joint
(i.e. all per-exit losses permuted together) vs. per-exit permutation-invariance where
speakers can swap between exits, (d) and (e) the number of exits --- we train 4-, 6-,
and 12-exit variants of our small model configuration with exit-points placed uniformly
over depth, and (f) where we ablate that simply finetuning on more 4-second audio does
not lead to the same performance improvements as training on full-length data.

Ablation results can be found in~\cref{table:result_ablations}. For ablation (a) we see
that our Student t-likelihood can be used in place of \ac{sisnr} without loss of
performance, even though our t-likelihood is not scale-invariant. Ablation (b) shows
that a simpler normal likelihood results in worse reconstruction performance in terms
of \ac{sisnr}, possibly due to not log-scaling the error as both \ac{sisnr} and the
t-likelihood do. Ablation (c) reveals that permuting consecutive early exits together
is crucial for stable joint training of early exits, likely because per-exit
permutation-invariance would allow the network to swap sources through the speaker
attention layers, defeating much of the point of the early split architecture design.
In ablations (d) and (e) we found that increasing the number of exits from 4 to 6 or 12
did not worsen performance at any of the exit points, which motivated us to train
larger models at the 12-exit level. We also see in ablation (f) that additional
finetuning with 4-second audio does not increase performance.

\begin{figure}[t]
  \centering
  \includegraphics[width=\linewidth]{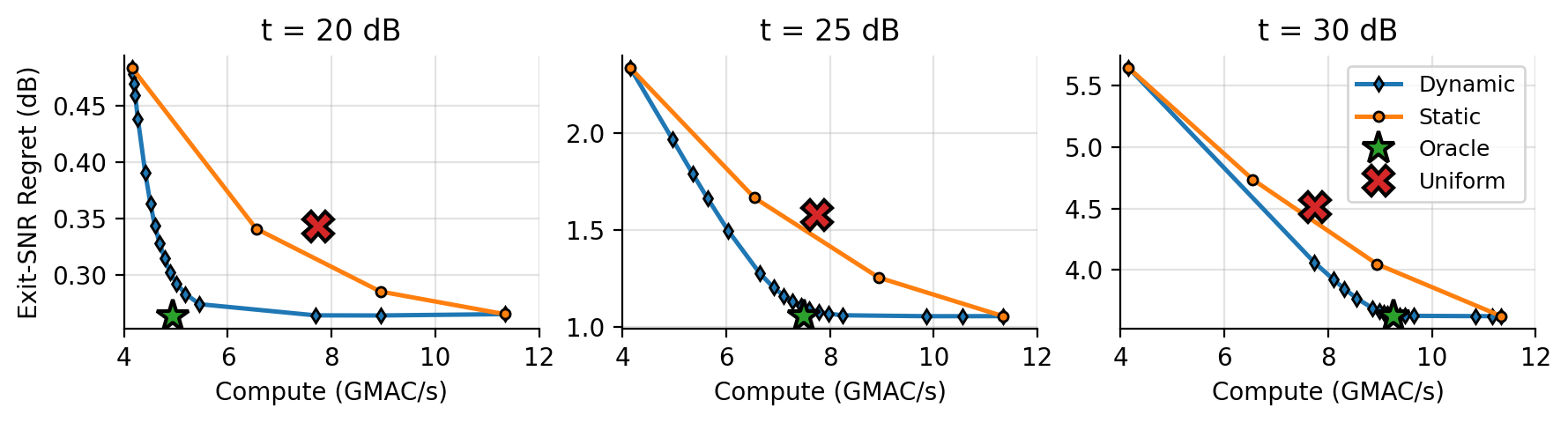}
  \caption[hmm]{
    \textbf{One-sided exit-SNR regret} on the WSJ0-2mix test set for a PRESS-4 (S) model
    trained with a block size of 2000 samples using different early-exit
    strategies with target levels of $t=20,25,30$ dB:
    \textbf{(dynamic)} our probabilistic exit strategy in~\cref{eq:exit_snr}
    evaluated for varying confidence thresholds $p$, \textbf{(static)}
    using a single exit for all blocks, \textbf{(oracle)} a best-case strategy that
    always exits when the target is achieved using the ground-truth exit-SNR,
    \textbf{(uniform)} an uninformed strategy that selects exit points uniformly at
    random.
  } \label{fig:regret_pareto}
\end{figure}

We further investigate the use of the \ac{this} architecture the DNS2020 dataset by
treating the speech enhancement task as a source separation, where we predict both the
clean speech and noise signals as separate sources. Surprisingly, as seen
in~\cref{table:result_comparison_dns}, this leads to very competitive performance after
accounting for total \ac{gmacs} even though our model also explicitly recovers the
noise signal while other methods do not, with our smaller PRESS-12 (M) model
configuration matching the ZipEnhancer~\autocite{WangZipenhancer2025} method in terms
of \ac{sisnri} using substantially less compute.

We also experiment with evaluating our exit-SNR condition in~\cref{fig:regret_pareto},
at 3 different target levels by varying the confidence threshold $p$, and measuring the
one-sided regret of our exit condition, i.e. the difference between the achieved
exit-SNR and the target level, or 0 if the exit-SNR exceeded the target. We see that
using our early exit strategy closely matches the oracle strategy at the appropriate
confidence level.

\begin{table}
  \centering
  \fontsize{8}{8}\selectfont
  \renewcommand{\arraystretch}{1.3}
  \sisetup{
    table-number-alignment = center,
    table-format = 2.2(2),
    detect-all = true,
    separate-uncertainty = true
  }
  \begin{tabular}{
      l
      S[table-format=2.2]
      S[table-format=2.2]
      S[table-format=1.2]
      S[table-format=2.2]
      S[table-format=3.1]
    }
     & \textbf{SI-SDR} & \textbf{STOI} & \textbf{WB-PESQ} & \textbf{\# Params} & \textbf{GMAC/s} \\
    \toprule
    MFNet~\autocite{LiuAMaskFree2023}
     & 20.31           & 97.98         & 3.43             & \textendash        & 6.1             \\
    MP-SENet~\autocite{LuMpSenet2023}
     & 21.03           & 98.16         & 3.62             & 2.26M              & 40.7            \\
    TF-Locoformer~\autocite{SaijoTfLocoformer2024}
     & 23.30           & 98.80         & 3.72             & 14.97M             & 248.6           \\
    ZipEnhancer~\autocite{WangZipenhancer2025}
     & 22.22           & 98.65         & 3.81             & 11.34M             & 133.5           \\
    \midrule
    \rowcolor{lightgray}
    PRESS-4 @ 4 (S)
     & 20.53           & 96.34         & 2.69             & 3.55M              & 11.6            \\
    \rowcolor{lightgray}
    PRESS-12 @ 4 (M)
     & 20.97           & 96.52         & 2.92             & 8.57M              & 29.1            \\
    \rowcolor{lightgray}
    PRESS-12 @ 8 (M)
     & 21.98           & 96.97         & 3.10             & 14.95M             & 53.7            \\
    \rowcolor{lightgray}
    PRESS-12 @ 12 (M)
     & 22.15           & 97.13         & 3.10             & 18.14M             & 78.3            \\
    \bottomrule
  \end{tabular}
  \caption[hmm]{
    \textbf{Speech enhancement performance} on the DNS2020 non-blind test set without
    reverberation.
  }\label{table:result_comparison_dns}
\end{table}

\section{Conclusion \& Future Work}

We introduced the \ac{this} method and \ac{this}-Net model, achieving competitive
performance on WSJ0-2Mix, Libri2Mix, WHAM!, WHAMR!, and DNS Challenge 2020 while
allowing flexible compute scaling based on probabilistic early-exit conditions.

Our probabilistic approach allows well-defined early exit conditions to be formulated
with integrated uncertainty quantification using the CDF of the exit-SNR distributions,
and can be used in place of conventional training objectives such as \ac{sisnr} at no
apparent cost to the reconstruction objective.

Our predictive \ac{snr}-like distributions proved to be very well-calibrated after
finetuning on full-length data, which also provided significant performance
improvements to reconstruction.

An extension of our work would be to consider iterative models, i.e. special cases of
our model with a single shared block in the decoder stack repeated for each exit point.
This would allow theoretically infinite scaling with compute, but if done naively
couples the total parameter count to the size of the iterative block requiring more
careful network design, possibly using width-scaling neural networks.

\newpage

\section*{Acknowledgements}

This work is partly funded by WS Audiology and the Innovation Fund Denmark (IFD) under
File No.\ 3129-00075B.

\bibliography{refs.bib}

\newpage

\appendix

{\LARGE\sc Appendix}

\section{$\chi^2$-Ratio Approximation}\label{sec:chisquare}

In~\cref{sec:prob_separation} we claim that our model assumptions allow us to express a
probabilistic \ac{snri} in terms of the estimated error variance $\sigma^2$. From our
likelihood we have (omitting speaker indices $i$ and $j$),
\begin{equation}
  \vec{x} \sim \gaussian{\vec{\widehat{x}}, \sigma^2 \mat{I}}, \qquad
  \sigma^2 \sim \invgammadist{\alpha, \beta}.
\end{equation}
which we can equivalently write as,
\begin{equation}
  \vec{x} = \vec{\widehat{x}} + \sigma \vec{z}, \qquad \vec{z} \sim
  \gaussian{\vec{0}, \mat{I}}.
\end{equation}
We then obtain
\begin{gather}
  \phi = \norm{\vec{x} - \vec{\widetilde{x}}}_2^2 = \norm{\vec{\widehat{x}} +
  \sigma \vec{z} - \vec{\widetilde{x}}}_2^2 = \sigma^2 \norm{\vec{z} +
  \frac{\vec{\widehat{x}} - \vec{\widetilde{x}}}{\sigma}}_2^2 \sim \sigma^2
  \chisq{T}{\frac{\norm{\vec{\widehat{x}} - \vec{\widetilde{x}}}_2^2}{\sigma^2}}, \\
  \epsilon = \norm{\vec{x} - \vec{\widehat{x}}}_2^2 = \sigma^2
  \norm{\vec{z}}_2^2 \sim \sigma^2 \chi^2_T, \\
  \mathrm{SNRi} = \frac{\phi}{\epsilon}.
\end{gather}
These distributions are dependent on the same draw of $\vec{z}$ and consequently the
distribution of $\frac{\phi}{\epsilon}$ does \emph{not} take the form of a non-central
F-distribution, nor can we rely on existing dependence results that assume shared
sub-summations in the two associated $\chi^2$
distributions~\autocite{ProvostTheExactDensity1994} or the correlation to be
known~\autocite{JoarderMomentsOfThe2009}.

Notably, the $\sigma^2$ scaling factors cancel in the ratio, so that the only sources
of randomness are (1) the draw of $\sigma^2$ in the non-centrality parameter of $\phi$
and (2) the draw of $\vec{z}$. We now instead consider the limiting behavior of the
ratio as $T \to \infty$ to see how it may be approximated for large $T$. Since
$\norm{\vec{\widehat{x}} - \vec{\widetilde{x}}}_2^2$ is a constant which scales
linearly with $T$, we can rewrite the non-centrality parameter of $\phi$ to $T
\lambda$, where we have absorbed the constant scale into a new random variable $\lambda
= \frac{\norm{\vec{\widehat{x}} - \vec{\widetilde{x}}}_2^2}{T \sigma^2} \sim
\gammadist{\alpha, \frac{\norm{\vec{\widehat{x}} - \vec{\widetilde{x}}}_2^2}{T
\beta}}$. We then have to show that for
\begin{equation}
  \phi = X_T \sim \chisq{T}{T \lambda}, \qquad \epsilon = Y_T \sim \chi^2_T,
\end{equation}
the ratio $\frac{X_T}{Y_T}$ converges in distribution to
\begin{equation}
  \frac{X_T}{Y_T} \xrightarrow{d} 1 + \lambda \quad \text{as} \quad T \to \infty.
\end{equation}
We start by noting that the normalized variables $\frac{X_T}{T}$ and $\frac{Y_T}{T}$
have decreasing variance as $T \to \infty$,
\begin{gather}
  \expect*{\frac{X_T}{T} \given \lambda} = \frac{\expect*{X_T \given
  \lambda}}{T} = 1 + \lambda, \qquad \var*{\frac{X_T}{T} \given \lambda} =
  \frac{\var*{X_T \given \lambda}}{T^2} = \frac{2 + 4\lambda}{T} =
  \bigO*{\frac{1}{T}}, \\
  \expect*{\frac{Y_T}{T}} = \frac{\expect*{Y_T}}{T} = 1, \qquad
  \var*{\frac{Y_T}{T}} = \frac{\var*{Y_T}}{T^2} = \frac{2}{T} = \bigO*{\frac{1}{T}},
\end{gather}
which implies that both normalized variables converge in probability to their expected
values as $T \to \infty$. By Slutsky's
theorem~\autocite{CasellaStatisticalInference2024}, the ratio $\frac{X_T}{Y_T}$ then
also converges in distribution as,%
\begin{equation}
  \frac{X_T}{Y_T} = \frac{\frac{X_T}{T}}{\frac{Y_T}{T}} \xrightarrow{d}
  \frac{\expect*{\frac{X_T}{T} \given \lambda}}{\expect*{\frac{Y_T}{T}}} = 1
  + \lambda \quad \text{as} \quad T \to \infty.
\end{equation}
This convergence holds even when $X_T$ and $Y_T$ are not independent, as the normalized
variables converge to the expected values separately. The \ac{snr} case is analogous,
substituting $\norm{\vec{\widehat{x}} - \vec{\widetilde{x}}}$ for
$\norm{\vec{\widehat{x}}}$.

\begin{figure}[htpb]
  \centering
  \includegraphics[width=0.5\linewidth]{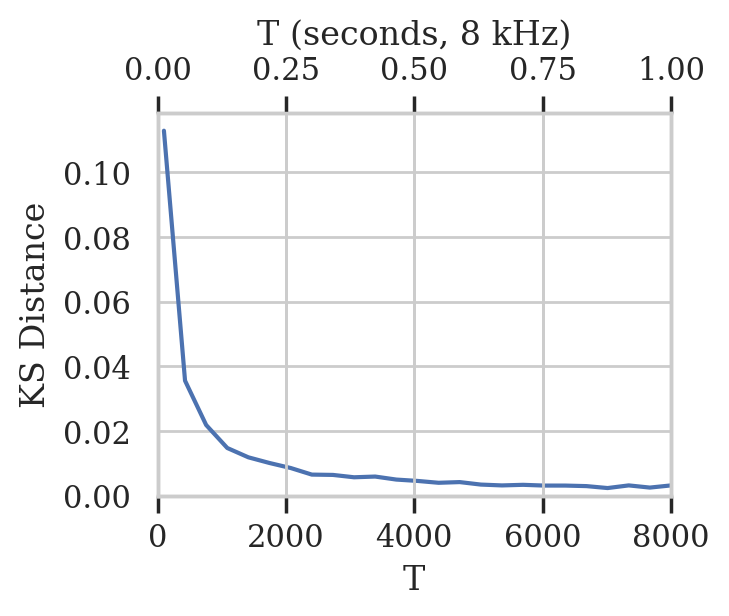}
  \caption{
    \textbf{Simulated \ac{ks} distances} between the simulated and approximate
    \ac{snri} distributions for varying values of $T$. The \ac{ks} statistic
    represents the maximum deviation between the empirical and limiting CDFs,
    which is below 1\% after $T \approx 2000$.
  } \label{fig:kstest}
\end{figure}

To get a feeling for the rate of convergence, we simulate the true distribution of
$\frac{X_T}{Y_T}$ for finite $T$ with $\lambda \sim \gammadist{\alpha,
\frac{c}{\beta}}$ and perform one-sample Kolmogorov-Smirnov tests comparing the
empirical distribution of the ratio for $1\,000\,000$ samples to the limiting
distribution $1 + \lambda$ in~\cref{fig:kstest}. We set the parameters $\alpha=60$,
$\beta=0.1$, and $c=0.1$ based on the average model predictions seen during training.

In~\cref{tab:ks_values} we show the average \ac{ks} distance between approximate and
simulated \ac{snri} distributions on the WSJ0-2mix test set for our blocked-likelihood
PRESS-4 (S) models, demonstrating that in practice even relatively small values of $T$
are still approximated well by our asymptotic distribution.

\begin{table}[htpb]
  \centering
  \sisetup{
    table-number-alignment = center,
    table-format = 2.2(2),
    detect-all = true,
    separate-uncertainty = true
  }
  \begin{tabular}{
      l
      S[table-format=1.3]
      S[table-format=4]
    }
    {\textbf{Ablation}} & {\textbf{\Ac{ks} distance}} & {\textbf{Receptive field}} \\
    \toprule
    Block size $T=8000$ & 0.007                       & 1000 ms                    \\
    Block size $T=4000$ & 0.008                       & 500 ms                     \\
    Block size $T=2000$ & 0.012                       & 250 ms                     \\
    Block size $T=1000$ & 0.019                       & 125 ms                     \\
    Block size $T=500$  & 0.049                       & 62 ms                      \\
    \bottomrule
  \end{tabular}
  \caption[hmm]{
    \textbf{Test \ac{ks} distances} on the WSJ0-2mix test set.
  } \label{tab:ks_values}
\end{table}

\section{Full Early Exit Distributions}\label{sec:full_demo}

\begin{figure}[!t]
  \centering
  \includegraphics[width=\linewidth]{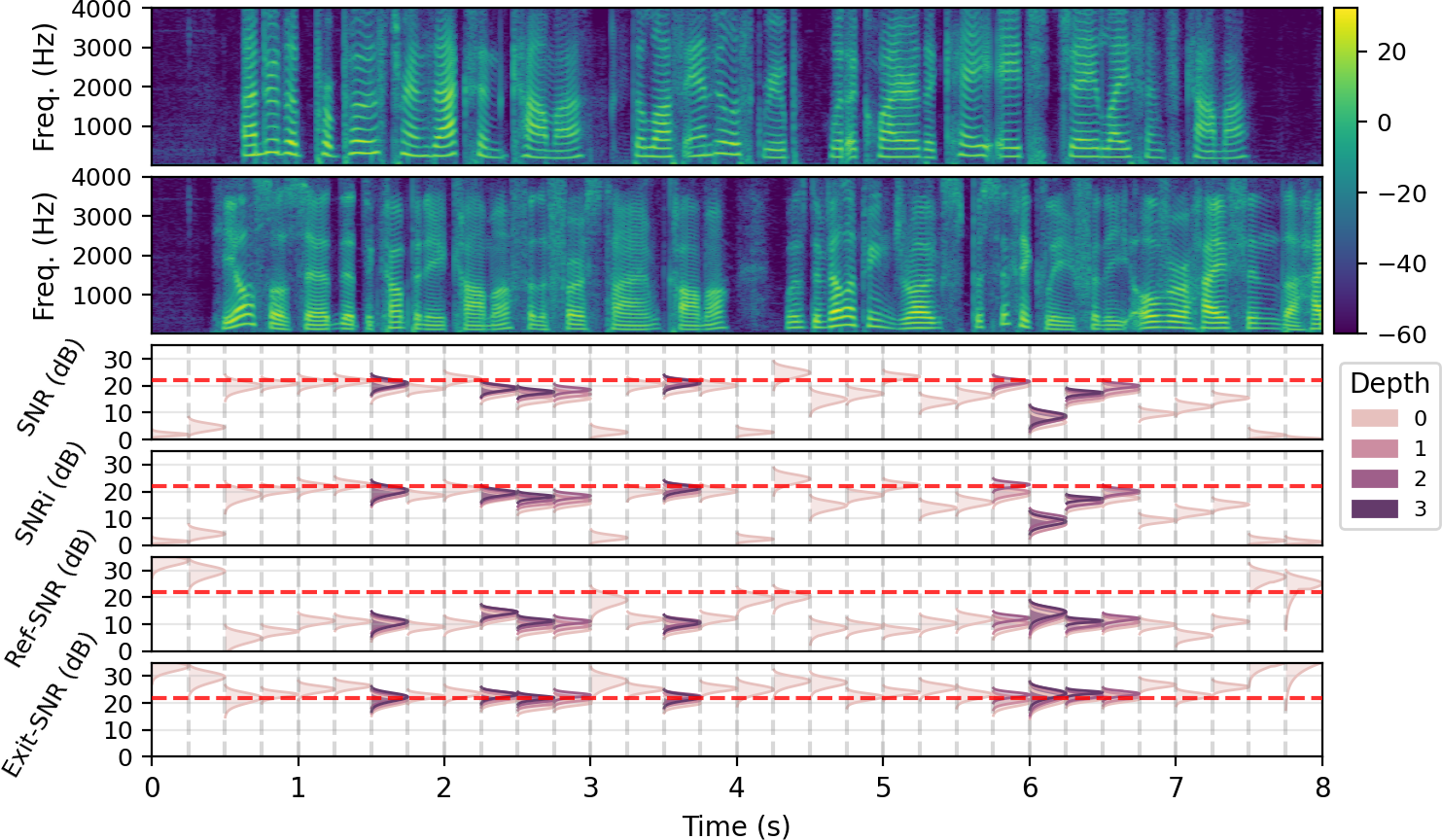}
  \caption{
    Full version of~\cref{fig:early_exit_demo} including separate exit conditions.
  }
  \label{fig:early_exit_demo_full}
\end{figure}

A complete version of~\cref{fig:early_exit_demo} can be seen
in~\cref{fig:early_exit_demo_full}.

\section{Model Architecture Details}\label{sec:arch_details}

\begin{figure}[htpb]
  \centering
  \hfill
  \begin{subfigure}{0.26\textwidth}
    \includegraphics[width=\textwidth]{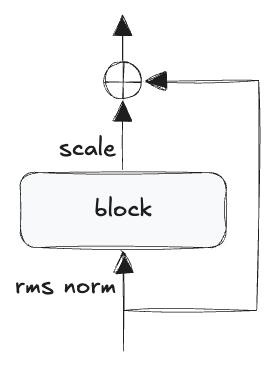}
    \caption{
    } \label{fig:arch:structure}
  \end{subfigure}
  \hfill
  \begin{subfigure}{0.3\textwidth}
    \includegraphics[width=\textwidth]{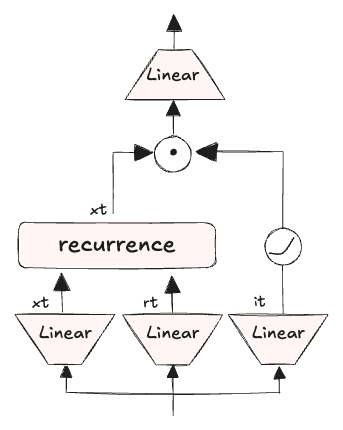}
    \caption{
    } \label{fig:arch:temporal}
  \end{subfigure}
  \hfill
  \begin{subfigure}{0.26\textwidth}
    \includegraphics[width=\textwidth]{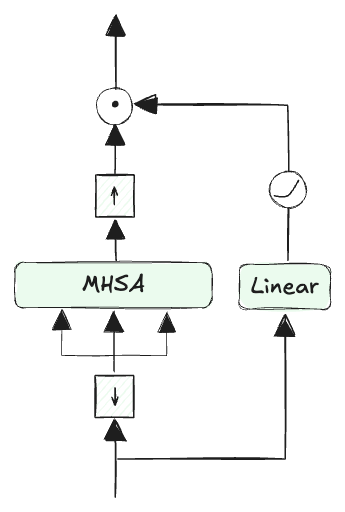}
    \caption{
    } \label{fig:arch:conv}
  \end{subfigure}
  \hfill
  \begin{subfigure}{0.18\textwidth}
    \includegraphics[width=\textwidth]{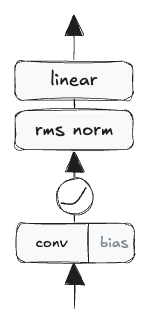}
    \caption{
    } \label{fig:arch:enchead}
  \end{subfigure}
  \hfill
  \begin{subfigure}{0.30\textwidth}
    \includegraphics[width=\textwidth]{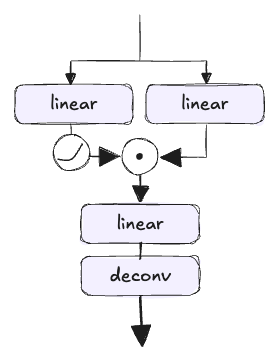}
    \caption{
    } \label{fig:arch:dechead}
  \end{subfigure}
  \hfill
  \begin{subfigure}{0.30\textwidth}
    \includegraphics[width=\textwidth]{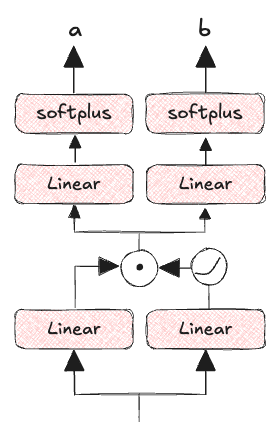}
    \caption{
    } \label{fig:arch:invgamma}
  \end{subfigure}
  \hfill
  \caption{
    Detailed architecture of the blocks used. \textbf{(a)} shows the overall block
    architecture where the \textit{block} which may be either \textbf{(b)} the linear
    \ac{rnn} or \textbf{(c)} speaker attention blocks. \textbf{(d)} and
    \textbf{(e)} are the audio encoder and decoder heads respectively and
    \textbf{(f)} is the InvGamma parametrization block.
  } \label{fig:arch:blocks}
\end{figure}

\paragraph{Audio Encoder and Decoder Heads} We base our encoder and decoder
heads on the design from SepReformer, with the audio encoder head processing
the time-domain audio signal $\vec{x}$ as $\mathbb{R}^T \to \mathbb{R}^{D_{enc}
\times T / P} $ by passing it through a 1-D convolution with kernel size
$K=16$, stride $P=4$ and encoding dimension $D_\text{enc}=256$ with bias, which
is used to avoid division-by-zero in later RMSNorm layers that would otherwise
occur if an all-zero input signal was passed into the model. The representation
is then passed through a GELU, RMSNorm and finally a linear layer which
projects down to the model dimension $D_\text{model}=64 \text{ or } 128$.
See~\cref{fig:arch:enchead}.

The decoder head consists of a \ac{glu} layer followed by a transposed convolution with
kernel size $K=16$, and stride $P=4$ that maps the latent representation back to the
original sampling rate as $\mathbb{R}^{D_{model} \times T/P} \to \mathbb{R}^{T}$. Every
exit point has its own decoder head; see~\cref{fig:arch:dechead}.

\paragraph{Linear RNNs} To capture rich, long-range temporal relationships
without the quadratic cost of attention~\autocite{VaswaniAttentionIsAll2023},
we use a linear \ac{rnn} based on minGRU~\autocite{FengWereRnnsAll2024} and
RG-LRU~\autocite{DeGriffin2024}. The recurrence $\operatorname{R}(\cdot,
\cdot)$ takes as input $x_t \in \mathbb{R}^D$ and $r_t \in \mathbb{R}^D$ to
produce an output sequence $h_t \in \mathbb{R}^D$ as follows:
\begin{equation}
  h_t = g_t \odot h_{t-1} + (1-g_t) \odot x_t, \qquad g_t =
  \sigma(\lambda)^{\sigma(r_t)},\label{eq:recurrence}
\end{equation}
where $\lambda$ is a learnable diagonal matrix, and $\operatorname{\sigma}(\cdot)$ is
the sigmoid function. This parametrization ensures that $g_t$ is bounded between 0 and
1, making the recurrence stable. $\lambda$ is initialized such that $\sigma(\lambda)$
is uniformly distributed between 0.9 and 0.999 as in~\textcite{DeGriffin2024}.

Notice that $\lambda$ being a diagonal matrix, makes the computation of $g_t$
elementwise. Moreover, the recurrence itself operates entirely in an elementwise
fashion. Since the recurrence is linear in terms of $h_t$ and since $g_t$ does not
depend on $h_t$, the recurrence can be parallelized along the time dimension using a
parallel associative
scan~\autocite{BlellochPrefixSumsAnd1990,MartinParallelizingLinearRecurrent2018} for
efficient training.

We use the quasiseparable matrix framework from~\textcite{HwangHydra2024}, which we
refer to as just Hydra bidirectionality, in order to construct a bidirectional variant
of our recurrence:
\begin{equation}
  \operatorname{Hydra}_{\operatorname{R}}(r_t, x_t) =
  \operatorname{shift}(\operatorname{R}(r_t, x_t)) +
  \operatorname{flip}(\operatorname{shift}(\operatorname{R}(\operatorname{flip}(r_t),
  \operatorname{flip}(x_t)))),
\end{equation}
where $\operatorname{flip}(\cdot)$ reverses a sequence along the time dimension and
$\operatorname{shift}(\cdot)$ shifts the sequence one time index to the right,
discarding the final time step and prepending the sequence with a zero vector $0 \in
R^{D}$. Notice that $r_t$ and $r_t$ are shared in both terms of the bidirectionality,
such that the Hydra approach does not require additional learned parameters. This
bidirectionality has been shown to give better performance than simpler alternatives
like additive or concatenative approaches~\autocite{HwangHydra2024}.

We use linear RNNs in a self-gating block that projects the input into $x_t$ and $r_t$
for use in~\cref{eq:recurrence}, applies the recurrence, multiplies the output with a
multiplicative branch gated by a GELU activation, before finally projecting the output
with another linear layer. This block closely resembles the block used in
Mamba~\autocite{GuMamba2023}, but without convolutions which we found did not affect
performance. See~\cref{fig:arch:temporal}.

\paragraph{Inverse Gamma Parametrization} We model the parameters $\alpha$
and $\beta$ of the inverse-gamma distribution
in~\cref{eq:studentt_likelihood} with the InvGam block
shown in~\cref{fig:arch:invgamma}. The
block is composed of a \ac{glu} layer followed by a GELU activation, which is then
linearly projected to two scalars followed by a final softplus activation to the
parameters to be non-negative.

\paragraph{Block-wise Early-Exits} Early-exiting with our block-wise likelihood
requires managing RNN state and reconstruction from overlapping transposed
convolutions in the decoder heads. We handle RNN states by initializing all
states to zero whenever an RNN layer is ``restarted'' after an earlier exit, and we
combine contributions from different exit points across block boundaries by
overlap-adding the transposed convolutions from separate layers, corresponding
to smoothly interpolating between partial results from both contributions.

\section{Dataset Descriptions}\label{sec:dataset_descriptions}

\paragraph{WSJ0-2mix} The WSJ0-2mix~\autocite{GarofoloCsrIComplete2007}
dataset is generated according to a
Python replica\footnote{\url{https://github.com/mpariente/pywsj0-mix}} of the original
Matlab implementation. 2-speaker mixtures are generated by randomly selecting
utterances from different speakers in the Wall Street Journal (WSJ0) corpus. The
training set contains 20,000 mixtures (30 hours), the validation set has 5,000 mixtures
(8 hours), and the test set comprises 3,000 mixtures (5 hours). Speech mixtures are
created by mixing pairs of utterances from different speakers at random \acp{snr}
drawn uniformly between 0 and 5 dB.

\paragraph{Libri2Mix} The Libri2Mix~\autocite{CosentinoLibrimix2020} dataset
is generated according to the
official script\footnote{\url{https://github.com/JorisCos/LibriMix}}, which creates
2-speaker mixtures by randomly selecting utterances from different speakers in the
LibriSpeech corpus. The loudness of each individual utterance is uniformly sampled
between -25 and -33 LUFS. This mixing approach results in signal-to-noise ratios (SNRs)
that follow a normal distribution with a mean of 0 dB and a standard deviation of 4.1
dB. Only the train-100set, which has 40hours/9300utterances of data, is used for
training and results are reported on the test set which contains 3000 samples.

\paragraph{WHAM!} The WHAM!~\autocite{WichernWham2019} dataset extends the WSJ0-2mix
mixtures with additive environmental noise recorded in various urban
environments. The mixtures are generated following the same procedure as for
WSJ0-2mix, but with noise added at an \ac{snr} sampled uniformly in the -6 and
3 dB range between the loudest speaker and noise.

\paragraph{WHAMR!} The WHAMR!~\autocite{MaciejewskiWhamr2020} dataset in turn
extends WHAM! by introducing reverberation into the WSJ0-2mix speech mixtures
before adding the WHAM! noise by convolving the speech signals with
artificially generated room impulse responses designed to approximate domestic
and class-room environments.

\paragraph{DNS2020} The Deep Noise Suppression (DNS) Challenge
2020~\autocite{ReddyTheInterspeechDeep2020} dataset includes two main parts: 441
hours of clean speech extracted from LibriVox~\autocite{NoauthorLibrivox2005}
audiobooks and a noise library of about 195 hours. The noise library combines
60,000 clips from AudioSet with 10,000 clips from
Freesound~\autocite{FontFreesoundTechnicalDemo2013} and
DEMAND~\autocite{ThiemannTheDiverseEnvironments2013} datasets. DNS2020 provides
a mixing script but sets no guidelines for how much data should be mixed.
Therefore, speech-noise mixtures were synthesized on the fly during training at
SNRs between 0-20 dB. This synthesis followed the procedure
provided in the original
script\footnote{\url{https://github.com/microsoft/DNS-Challenge/tree/interspeech2020/master}}.
Results are reported on the non-blind test set without reverberation. The test
set contains 150 premixed samples.

\paragraph{Sampling Rate and Downsampling} The speech separation datasets
operate at 8kHz sampling rate, while DNS2020 uses a
16kHz sampling rate. This creates the only difference between our models. For 8kHz
speech separation, we use a downsampling factor of 4 in the encoder, while for
the 16kHz DNS2020 we use a factor of 8, resulting in the same amount of
compute per second.

\section{Training Details: Datasets, Resources and Software}\label{sec:training_details}

We use the AdamW~\cite{KingmaAdam2017,LoshchilovDecoupledWeightDecay2019} optimizer
with $\beta_1=0.9$ and $\beta_2 = 0.99$, and weight decay of $0.01$ which we apply to
only linear and convolutional layers. We use a base learning rate of $5 \cdot 10^{-4}$,
which was found for a $D=64$ model with 4 exit points, and we transfer this learning
rate to wider models (e.g. $D=128$) by a per-layer factor of $\frac{D_{old}}{D_{new}}$
as described in~\textcite{YangASpectralCondition2024}. We use a linear straight-to-zero
learning rate schedule as described in~\textcite{BergsmaStraightToZero2025} with a
5000-step linear warmup period. During training we clip the total gradient if its L2
norm exceeds 1. All weight matrices were initialized from a normal distribution
truncated at $3\sigma$ with standard deviations set according
to~\textcite{YangASpectralCondition2024}.

For all datasets the models train on 4-second segments while at evaluation time the
model process samples of varying length. The models train for up to 6 million steps
with a batch size of 1, amounting to 6,666 hours of training data exposure.

We trained all models on NVIDIA GPUs with Ampere architecture or higher, using any of
H100, A100, A40, A10, RTX 4090, and RTX 4070 Ti at either a university HPC cluster or
commercially available GPUS
with PyTorch~\autocite{PaszkePytorch2019} 2.7 using \texttt{torch.compile}. The PRESS-4
(S) configurations took around 2-3 days to train, while the PRESS-12 (M) configurations
took around 6 days to train.

\section{Scale-Invariant Student t-Likelihood}\label{sec:scale_invariance}

Similar to \ac{sisnr}~\autocite{RouxSdrHalfBaked2018}, we can construct a
scale-invariant version of our likelihood by introducing a scaling parameter on the
estimated signal $\vec{\widehat{x}}$ which yields a modified likelihood,
\begin{equation}
  \mathcal{L} = \studentt{\vec{x} \given \gamma \vec{\widehat{x}}, 2\alpha,
  \frac{\beta}{\alpha}}.
\end{equation}
Any conjugate prior can be imposed on $\gamma$, but its effect would quickly be
overwhelmed by the posterior for non-trivial signal length. The maximum-likelihood
estimate for $\gamma$ yields,
\begin{equation}
  \gamma = \frac{\vec{\widehat{x}}^\top \vec{{x}}}{\vec{\widehat{x}}^\top
  \vec{\widehat{x}}},
\end{equation}
which closely resembles the scale-invariance parameter in the conventional SI-SNR, but
normalized by the energy in the prediction rather than the target. This
scale-invariance formulation has also been used with SI-SNR to train
TF-GridNet~\autocite{WangTfGridnet2023}.

Empirical studies show that \acp{tasnet} trained with log-RMS error perform comparably
to those trained with SI-SDR~\autocite{HeitkaemperDemystifyingTasnet2020}. While SI-SDR
losses yield better models than RMS error, the key factor may be the logarithmic error
scaling rather than scale invariance. Notably, the multivariate Student t-distribution
likelihood in~\cref{eq:studentt_likelihood} also measures errors on a log scale.

\end{document}